\documentclass[journal]{IEEEtran}
\IEEEoverridecommandlockouts
\usepackage{soul}
\usepackage{cite}
\usepackage{amsmath,amssymb,amsfonts}
\usepackage{algorithmic}
\usepackage{graphicx}
\usepackage{textcomp}
\usepackage{bm,upgreek}
\usepackage[dvipsnames]{xcolor}
\usepackage[caption=false,labelformat=simple]{subfig}

\usepackage[left=0.68in, right=0.68in, bottom=0.351in, top=0.772in, includefoot]{geometry}

\DeclareMathOperator{\pred}{Pred}
\DeclareMathOperator{\layerv}{Layer^v}
\DeclareMathOperator{\layerf}{Layer^f}
\DeclareMathOperator{\messageFN}{Message}
\DeclareMathOperator{\updateFN}{Update}
\DeclareMathOperator{\aggregateFN}{Aggregate}

\def\BibTeX{{\rm B\kern-.05em{\sc i\kern-.025em b}\kern-.08em
    T\kern-.1667em\lower.7ex\hbox{E}\kern-.125emX}}

\usepackage{pgfplots}
\pgfplotsset{compat=1.5}
\usepackage{pgfplotstable}
\usepgfplotslibrary{groupplots}
\usepgfplotslibrary{statistics}
\pgfplotsset{grid style={dotted,gray}}
\usepackage{tikz}
\usepackage[toc,acronym]{glossaries}
\newacronym{3gpp}{3GPP}{Third Generation Partnership Project}
\newacronym{ran}{RAN}{Radio Access Network}
\newacronym{oran}{O-RAN}{Open Radio Access Network}
\newacronym{ml}{ML}{Machine Learning}
\newacronym{ai}{AI}{Artificial Intelligence}
\newacronym{sba}{SBA}{Service-Based Architecture}
\newacronym{5g}{5G}{Fifth-Generation}
\newacronym{4g}{4G}{Fourth-Generation}
\newacronym{cn}{CN}{Core Network}
\newacronym{ngran}{NG-RAN}{Next Generation Radio Access Network}
\newacronym{urllc}{URLLC}{Ultra-Reliable Low-Latency Communications}
\newacronym{nf}{NFs}{Network Functions}
\newacronym{nef}{NEF}{Network Exposure Function}
\newacronym{nfv}{NFV}{Network Functions Virtualization}
\newacronym{sdn}{SDN}{Software-Defined Networking}
\newacronym{sg}{SG}{Smart Grid}
\newacronym{pcf}{PCF}{Policy Control Function}
\newacronym{iot}{IoT}{Internet of Things}
\newacronym{se}{SE}{State Estimation}
\newacronym{ems}{EMS}{Energy Management System}

\pgfplotsset{legend image with text/.style={legend image code/.code={%
\node[anchor=west, align=right] at (0.0cm,0cm) {#1};}},}

\pgfplotsset{
    box plot/.style={
        /pgfplots/.cd,
        black,
        only marks,
        mark=-,
        mark size=\pgfkeysvalueof{/pgfplots/box plot width},
        /pgfplots/error bars/y dir=plus,
        /pgfplots/error bars/y explicit,
        /pgfplots/table/x index=\pgfkeysvalueof{/pgfplots/box plot x index},
    },
    box plot box/.style={
        /pgfplots/error bars/draw error bar/.code 2 args={%
            \draw  ##1 -- ++(\pgfkeysvalueof{/pgfplots/box plot width},0pt) |- ##2 -- ++(-\pgfkeysvalueof{/pgfplots/box plot width},0pt) |- ##1 -- cycle;
        },
        /pgfplots/table/.cd,
        y index=\pgfkeysvalueof{/pgfplots/box plot box top index},
        y error expr={
            \thisrowno{\pgfkeysvalueof{/pgfplots/box plot box bottom index}}
            - \thisrowno{\pgfkeysvalueof{/pgfplots/box plot box top index}}
        },
        /pgfplots/box plot
    },
    box plot top whisker/.style={
        /pgfplots/error bars/draw error bar/.code 2 args={%
            \pgfkeysgetvalue{/pgfplots/error bars/error mark}%
            {\pgfplotserrorbarsmark}%
            \pgfkeysgetvalue{/pgfplots/error bars/error mark options}%
            {\pgfplotserrorbarsmarkopts}%
            \path ##1 -- ##2;
        },
        /pgfplots/table/.cd,
        y index=\pgfkeysvalueof{/pgfplots/box plot whisker top index},
        y error expr={
            \thisrowno{\pgfkeysvalueof{/pgfplots/box plot box top index}}
            - \thisrowno{\pgfkeysvalueof{/pgfplots/box plot whisker top index}}
        },
        /pgfplots/box plot
    },
    box plot bottom whisker/.style={
        /pgfplots/error bars/draw error bar/.code 2 args={%
            \pgfkeysgetvalue{/pgfplots/error bars/error mark}%
            {\pgfplotserrorbarsmark}%
            \pgfkeysgetvalue{/pgfplots/error bars/error mark options}%
            {\pgfplotserrorbarsmarkopts}%
            \path ##1 -- ##2;
        },
        /pgfplots/table/.cd,
        y index=\pgfkeysvalueof{/pgfplots/box plot whisker bottom index},
        y error expr={
            \thisrowno{\pgfkeysvalueof{/pgfplots/box plot box bottom index}}
            - \thisrowno{\pgfkeysvalueof{/pgfplots/box plot whisker bottom index}}
        },
        /pgfplots/box plot
    },
    box plot median/.style={
        /pgfplots/box plot,
        /pgfplots/table/y index=\pgfkeysvalueof{/pgfplots/box plot median index},
        semithick,black
    },
    box plot width/.initial=1em,
    box plot x index/.initial=0,
    box plot median index/.initial=1,
    box plot box top index/.initial=2,
    box plot box bottom index/.initial=3,
    box plot whisker top index/.initial=4,
    box plot whisker bottom index/.initial=5,
}

\newcommand{\boxplot}[2][]{
    \addplot [box plot median,#1] table {#2};
    \addplot [forget plot, box plot box,#1] table {#2};
    \addplot [forget plot, box plot top whisker,#1] table {#2};
    \addplot [forget plot, box plot bottom whisker,#1] table {#2};
}

\begin{document}

\title{Distributed Nonlinear State Estimation in Electric Power Systems using Graph Neural Networks}

\author{Ognjen~Kundacina,~\IEEEmembership{Student Member,~IEEE,}
        Mirsad~Cosovic,~\IEEEmembership{Member,~IEEE,}
        Dragisa Miskovic,
        Dejan Vukobratovic,~\IEEEmembership{Senior Member,~IEEE}

\thanks{O. Kundacina and D. Miskovic are with The Institute for Artificial Intelligence Research and Development of Serbia (e-mail: ognjen.kundacina@ivi.ac.rs, dragisa.miskovic@ivi.ac.rs); 
 M. Cosovic is with Faculty of Electrical Engineering, University of Sarajevo, Bosnia and Herzegovina (e-mail: mcosovic@etf.unsa.ba); D. Vukobratovic is with Faculty of Technical Sciences, University of Novi Sad, Serbia, (email: dejanv@uns.ac.rs).}}

\IEEEoverridecommandlockouts
\IEEEpubid{\makebox[\columnwidth]{978-1-6654-3254-2/22/\$31.00~\copyright2022 IEEE \hfill} \hspace{\columnsep}\makebox[\columnwidth]{ }}

\maketitle

\begin{abstract}
Nonlinear state estimation (SE), with the goal of estimating complex bus voltages based on all types of measurements available in the power system, is usually solved using the iterative Gauss-Newton (GN) method. The nonlinear SE presents some difficulties when considering inputs from both phasor measurement units and supervisory control and data acquisition system. These include numerical instabilities, convergence time depending on the starting point of the iterative method, and the quadratic computational complexity of a single iteration regarding the number of state variables. This paper introduces an original graph neural network based SE implementation over the augmented factor graph of the nonlinear power system SE, capable of incorporating measurements on both branches and buses, as well as both phasor and legacy measurements. The proposed regression model has linear computational complexity during the inference time once trained, with a possibility of distributed implementation. Since the method is noniterative and non-matrix-based, it is resilient to the problems that the GN solver is prone to. Aside from prediction accuracy on the test set, the proposed model demonstrates robustness when simulating cyber attacks and unobservable scenarios due to communication irregularities. In those cases, prediction errors are sustained locally, with no effect on the rest of the power system's results.

\end{abstract}


\section{Introduction}
\textbf{Motivation:} The power system state estimation (SE) is a problem of determining the state of the power system represented as the set of complex bus voltages, given the available set of measurements \cite{monticelli2000SE}. The dominant part of the input data for the SE model consists of legacy measurements coming from the supervisory control and data acquisition (SCADA) system, which have relatively high variance, high latency, and low sampling rates. Increasingly deployed phasor measurement units (PMUs), provided by the wide area measurement system (WAMS), have low variance and high sampling rates and are a potential enabler of real-time system monitoring. Taking into account both legacy and phasor measurements results in the SE model formulated by the system of nonlinear equations and is traditionally solved using the iterative Gauss-Newton (GN) method. Different approaches can be used to integrate phasor measurements into the established model with legacy measurements. A standard way to include voltage and current phasors coming from PMUs is to represent them in the rectangular coordinate system \cite{catalina}. The main disadvantage of this approach is related to measurement errors, where measurement errors of a single PMU are correlated and covariance matrix does not have diagonal form. Despite that, because of the lower computational effort, the measurement error covariance matrix is usually considered as diagonal matrix, which has the effect on the accuracy of the SE. The diagonal form of the covariance matrix could be preserved by representing voltage and current phasors coming from PMUs in the polar coordinate system, which requires a large computational effort with a convergence time significantly depending on the state variables' initialisation \cite{manousakis}. Additionally, using magnitudes of branch current measurements can cause numerical instabilities such as undefined Jacobian elements due to the ``flat start" \cite[Sec. 9.3]{abur2004power}. Furthermore, different orders of magnitude of phasor and legacy measurement variances can make the SE problem ill-conditioned by increasing the condition number of the estimator's gain matrix \cite{catalina}. A single iteration of the GN method involves solving a system of linear equations, which results in near $\mathcal{O}(n^2)$ computational complexity for sparse matrices, where $n$ denotes the number of state variables.

Graph neural networks (GNNs) \cite{pmlr-v70-gilmer17a, GraphRepresentationLearningBook} are a promising method for iterative problems in large-scale systems because once trained, they have a $\mathcal{O}(n)$ computational complexity, with the possibility of distributed implementation. Apart from not being restricted to training and test examples with fixed topologies like the other deep learning approaches, GNNs are permutation invariant, have a smaller number of trainable parameters, require less memory for the deep learning model, and can easily incorporate topology information into the learning process. Since GNN inference is a noniterative procedure, its computational time depends on the power system size only and is not sensitive to state variable initialisation. Furthermore, it does not suffer from numerical instability since it is not based on the matrix model of the power system.

\textbf{Literature Review:} Various studies suggest using well-known deep learning models like feed-forward and recurrent neural networks to approximate the existing iterative SE solvers \cite{zhang2019} or to provide them with the state variable initialisation \cite{zamzam2019}. Recently, there have been several proposals for applying GNNs to similar power system analysis problems, like power flow \cite{donon2019graphneuralsolver, bolz2019PFapproximator} and probabilistic power flow \cite{wang2020ProbabPF}.

The work described in \cite{pagnier2021physicsinformed} presents a hybrid model and a data-based method for simultaneous power system parameters and state estimation using GNNs. The model predicts active and reactive power injections based on the bus voltage phasor measurements, whereas it does not take into account other measurement types, such as legacy measurements and branch current phasor measurements. The regularisation term in the SE loss function in \cite{Yang2022RobustPSSEDataDrivenPriors} uses a GNN trained on the historical dataset to forecast voltage labels. The GNN component of the proposed algorithm considers only inputs from node voltage measurements, while the other components consider different measurement types, resulting in an overall $\mathcal{O}(n^2)$ computational complexity during the inference phase. In \cite{TGCN_SE_2021}, state variables are predicted using a GNN with gated recurrent units trained on a time-series of node voltage measurements acquired from PMUs.

\textbf{Contributions:} In this work, we propose a data-driven nonlinear state estimator based on graph attention networks \cite{velickovic2018graph}. The proposed approach is an extension of our previous work on linear SE with PMUs \cite{kundacina2022state} and linear SE with PMUs considering covariances of measurement phasors represented in rectangular coordinates \cite{kundacina2022robust}. The contributions are listed as follows:
\begin{itemize}
    \item We provide an original implementation of GNNs over a factor graph \cite{Kschischang2001FactorGraphs} obtained by transforming the bus/branch power system model. Furthermore, we augment the factor graph to increase the information propagation in unobservable scenarios.
    \item The proposed method takes into account all of the legacy measurements, as well as bus voltage and branch current phasor measurements, and provides a trivial way to remove or add additional measurements by altering the corresponding factor nodes in the graph.
    \item We designed the GNN-based state estimator to be fast, robust to ill-conditioned scenarios, and an accurate nonlinear SE approximator, once trained on the relevant dataset, labelled by the nonlinear SE solved by GN. The proposed model has $\mathcal{O}(n)$ computational complexity during the inference and can be distributed among geographically separated processing units.
    \item We evaluated the performance of the proposed method by testing on various data samples, including unobservable cases caused by communication errors or measurement device failures, and scenarios corrupted by malicious data injections.
    \item In addition to the standalone application, the proposed approach can be used as a fast and accurate initialiser of the GN method by providing it with a starting point near the exact solution.
\end{itemize}

\section{Nonlinear State Estimation Preliminaries}
In a general scenario, the SE model is described with the system of nonlinear equations:
\begin{equation}
    \mathbf{z} = \mathbf{h}(\mathbf{x}) + \mathbf{u},
    \label{linear_model}
\end{equation}
where bus voltage magnitudes $\mathbf{V}$ and bus voltage angles $\bm \uptheta$ are observed as state variables $\mathbf{x} = [\mathbf{V}, \bm \uptheta]^T \in \mathbb{R}^{n}$, the vector $\mathbf{z} \in \mathbb{R}^{m}$ contains measurement values, and the vector $\mathbf{h}(\mathbf{x})$ comprises a collection of $m$ nonlinear equations. Next, $\mathbf{u} \in \mathbb{R}^{m}$ is a vector of uncorrelated measurement errors, where $u_i \sim \mathcal{N}(0, v_{i})$ represents a zero-mean Gaussian distribution with variance $v_{i}$. The nonlinear SE model \eqref{linear_model} is a consequence of legacy and phasor measurements provided by SCADA and WAMS, respectively. The typical set of legacy measurements provided by SCADA includes active and reactive power flow and injection, branch current magnitude, and bus voltage magnitude measurements. The WAMS supports phasor measurements and includes magnitude and angle measurements of bus voltages and branch currents \cite[Sec. 5.6]{phadke}.
   
Each measurement is associated with the measurement value $z_i$, the measurement variance $v_i$, and the measurement function $h_i(\mathbf{x})$. The form of the function $h_i(\mathbf{x})$ depends on the branch model of the power system. Usually, the two-port $\pi$-model is used, where the branch between buses $i$ and $j$ is settled using complex expressions:
\begin{equation}
  \begin{bmatrix}
    \bar{I}_{ij} \\ \bar{I}_{ji}
  \end{bmatrix} =
  \begin{bmatrix}
    \cfrac{1}{\tau_{ij}^2}(y_{ij} + y_{\text{s}ij}) & -\alpha_{ij}^*{y}_{ij}\\
    -\alpha_{ij}{y}_{ij} & {y}_{ij} + y_{\text{s}ij}
  \end{bmatrix}  
  \begin{bmatrix}
    \bar{V}_{i} \\ \bar{V}_{j}
  \end{bmatrix},
  \label{unified}
\end{equation} 
where the parameter $y_{ij} = g_{ij} + \text{j}b_{ij}$ represents the branch series admittance, and half of the total branch shunt admittance is given as $y_{\text{s}ij} = \text{j}b_{si}$. The transformer complex ratio is defined as $\alpha_{ij} = (1/\tau_{ij})\text{e}^{-\text{j}\phi_{ij}}$, where $\tau_{ij}$ is the transformer tap ratio magnitude, while $\phi_{ij}$ is the transformer phase shift angle. The complex expressions $\bar{I}_{ij}$ and $\bar{I}_{ji}$ define the branch currents from the bus $i$ to the bus $j$ and from the bus $j$ to the bus $i$, respectively. The complex bus voltages at buses $i$ and $j$ are given as $\bar{V}_i = V_i \text{e}^{\text{j}\theta_i}$ and $\bar{V}_j = V_j \text{e}^{\text{j}\theta_j}$, where $V_i$, $\theta_i$ and $V_j$, $\theta_j$ denote the voltage magnitude and angle at buses $i$ and $j$, respectively.

The GN method is typically used to solve the nonlinear SE model \eqref{linear_model}, where the measurement functions $\mathbf{h}(\mathbf{x})$ precisely follow the physical laws derived on the basis of \eqref{unified}:
		\begin{subequations}
        \begin{gather}  
		\Big[\mathbf J (\mathbf x^{(\nu)})^\mathrm{T} \mathbf W 
		\mathbf J (\mathbf x^{(\nu)})\Big] \Delta \mathbf x^{(\nu)} =
		\mathbf J (\mathbf x^{(\nu)})^\mathrm{T}
		\mathbf W\mathbf r (\mathbf x^{(\nu)})\label{AC_GN_increment}\\
		\mathbf x^{(\nu+1)} = 
		\mathbf x^{(\nu)} + \Delta \mathbf x^{(\nu)}, \label{AC_GN_update}
        \end{gather}
        \label{AC_GN}%
		\end{subequations}
where $\nu = \{0,1,\dots,\nu_{\max}\}$ is the iteration index and $\nu_{\max}$ is the number of iterations, $\Delta \mathbf x^{(\nu)} \in \mathbb {R}^{n}$ is the vector of increments of the state variables, $\mathbf J (\mathbf x^{(\nu)})\in \mathbb {R}^{m\mathrm{x}n}$ is the Jacobian matrix of measurement functions $\mathbf h (\mathbf x^{(\nu)})$ at $\mathbf x=\mathbf x^{(\nu)}$, $\mathbf{W}\in \mathbb {R}^{m\mathrm{x}m}$ is a diagonal matrix containing inverses of measurement variances, and $\mathbf r (\mathbf x^{(\nu)}) =$ $\mathbf{z}$ $-\mathbf h (\mathbf x^{(\nu)})$ is the vector of residuals. Note that the nonlinear SE represents a nonconvex problem arising from nonlinear measurement functions $\mathbf{h}(\mathbf{x})$ \cite{ilic}. Due to the fact that the values of the state variables $\mathbf{x}$ usually fluctuate in narrow boundaries, the GN method can be used.

The SE model \eqref{linear_model} that considers both legacy and phasor measurements, where the vector of state variables $\mathbf{x} = [\mathbf{V}, \bm \uptheta]^T$ and phasor measurements are represented in the polar coordinate system, is known as simultaneous. The simultaneous SE model takes measurements provided by PMUs in the same manner as legacy measurements. More precisely, the PMU generates measurements in the polar coordinate system, which delivers more accurate state estimates than the other representations \cite{catalina}, but requires more computing time \cite{manousakis} and produces ill-conditioned problems \cite{catalina}. To address these issues, we propose a non-matrix-based and noniterative GNN base SE, which can be used as a standalone approach to solve \eqref{linear_model}, or as a fast and accurate initialiser of the GN method \eqref{AC_GN}.

\section{Graph Neural Network based Nonlinear State Estimation}
This section introduces the fundamentals of spatial graph neural networks, the transformation of the power system's bus/branch model to an augmented factor graph, and the proposed GNN architecture, specialised for augmented factor graphs.

\subsection{Fundamentals of Spatial Graph Neural Networks} 

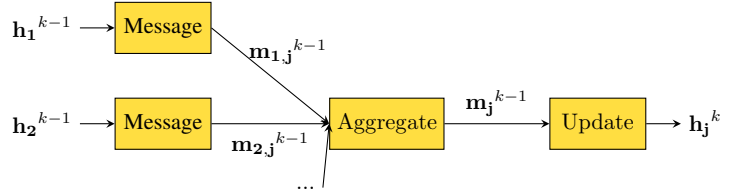
\begin{figure}[!t]
\centering

\begin{tikzpicture} [scale=0.85, transform shape]

\tikzset{
    box/.style={draw, fill=Goldenrod, minimum width=1.5cm, minimum height=0.8cm}}
    
\begin{scope}[local bounding box=graph]

\node [box]  (message1) at (-0.5, 1.5) {Message};
\node [box]  (message2) at (-0.5, 0) {Message};
\node [box]  (gat) at (3, 0) {$\aggregateFN$};
\node [box]  (update) at (6.3, 0) {$\updateFN$};


\draw[-stealth] (-1.8, 0) -- (message2.west) node[at start,left]{$\mathbf{h_{2}}^{k-1}$};

\draw[-stealth] (-1.8, 1.5) -- (message1.west) node[at start,left]{$\mathbf{h_{1}}^{k-1}$};

\draw[-stealth] (message1.east) -- (gat.west) node[near start,right]{$\mathbf{m_{1,j}}^{k-1}$};

\draw[-stealth] (message2.east) -- (gat.west) node[midway,below]{$\mathbf{m_{2,j}}^{k-1}$};
	
\draw[-stealth] (2, -1) -- (gat.west)
	node[at start,left]{$...$};
	
\draw[-stealth] (gat.east) -- (update.west) node[midway,above]{$\mathbf {m_j}^{k-1}$};

\draw[-stealth] (update.east) -- (7.6, 0) node[at end,right]{$\mathbf{h_{j}}^{k}$};


\end{scope}

\end{tikzpicture}

\caption{One message passing iteration, implemented as a GNN layer, consists of several trainable functions, shown as yellow rectangles with indicated inputs and outputs.}
    \label{GNNlayerDetails}
\end{figure}

The spatial GNNs perform recursive neighbourhood aggregation, also known as message passing \cite{pmlr-v70-gilmer17a}, over the local subsets of graph-structured inputs to create a meaningful representation of the connected pieces of data. More precisely, GNNs act as a trainable local graph filter which has a goal of transforming the inputs from each node and its connections to a higher dimensional space, resulting in an $s$-dimensional vector embedding $\mathbf h \in \mathbb {R}^{s}$ per node. The GNN layer, which represents one message passing iteration, combines multiple trainable functions commonly implemented as neural networks. The message function $\messageFN(\cdot): \mathbb {R}^{2s} \mapsto \mathbb {R}^{u}$ calculates the message $\mathbf m_{i,j} \in \mathbb {R}^{u}$ between two node embeddings, $\mathbf h_i$ and $\mathbf h_j$. The aggregation function $\aggregateFN(\cdot): \mathbb {R}^{\textrm{deg}(j) \cdot u} \mapsto \mathbb {R}^{u}$ combines the arriving messages in a specific way, and outputs the aggregated messages $\mathbf {m_j} \in \mathbb {R}^{u}$ for each node $j$. At the end of one message passing iteration, aggregated messages are fed into the update function $\updateFN(\cdot): \mathbb {R}^{u} \mapsto \mathbb {R}^{s}$, to calculate the update of each node's embedding. Node embedding values are initialised with the node's input feature vector transformed to the $s$-dimensional space, after which the message passing repeats $K$ times, with $K$ being a hyperparameter known as the number of GNN layers. One message passing iteration corresponding to the $k^{th}$ GNN layer is displayed in Fig.~\ref{GNNlayerDetails} and also described analytically by \eqref{gnn_equations}:
\begin{equation}
    \begin{gathered}
        \mathbf {m_{i,j}}^{k-1} = \messageFN( \mathbf {h_i}^{k-1}, \mathbf {h_j}^{k-1})\\
        \mathbf {m_j}^{k-1} = \aggregateFN(\{{\mathbf m_{i,j}}^{k-1} | i \in \mathcal{N}_j\})\\
        \mathbf {h_j}^k = \updateFN(\mathbf {m_j}^{k-1})\\
        k \in \{1,\dots,K\},
    \end{gathered}
    \label{gnn_equations}
\end{equation}
where $\mathcal{N}_j$ denotes the $1$-hop neighbourhood of the node $j$, and the vector superscript corresponds to the message passing iteration. The outputs of the message passing process are final node embeddings $\mathbf {h_j}^{K}$, which are then passed through the additional neural network layers, creating the outputs that represent the predictions of the supervised GNN model. GNN training is performed by optimising the model parameters using the variants of the gradient descent algorithm, with the loss function being some measure of the distance between the labels and the predictions. For a comprehensive overview of GNN foundations and algorithms, please refer to \cite{GraphRepresentationLearningBook}.

\subsection{Augmented Power System Factor Graph and the Proposed GNN Architecture} 
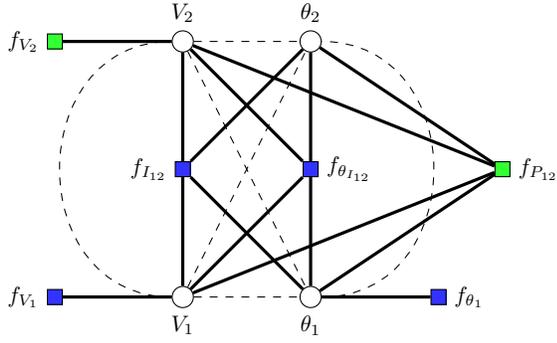
\begin{figure}[htbp]
    \centering
    \begin{tikzpicture} [scale=0.85, transform shape]
        \tikzset{
            varNode/.style={circle,minimum size=2mm,fill=white,draw=black},
            factorPMU/.style={draw=black,fill=blue!80, minimum size=2mm},
            factorSCADA/.style={draw=black,fill=green!80, minimum size=2mm},
            edge/.style={very thick,black},
            edge2/.style={dashed,black}}
        \begin{scope}[local bounding box=graph]
            \node[factorPMU, label=left:$f_{V_1}$] (f1) at (-3, 1 * 2) {};
            \node[factorPMU, label=right:$f_{\theta_1}$] (f4) at (3, 1 * 2) {};
            \node[varNode, label=below:$V_1$] (v1) at (-1, 1 * 2) {};
            \node[varNode, label=below:$\theta_1$] (v3) at (1, 1 * 2) {};
            \node[factorPMU, label=left:$f_{I_{12}}$] (f2) at (-1, 2 * 2) {};
            \node[factorPMU, label=right:$f_{\theta_{I_{12}}}$] (f5) at (1, 2 * 2) {};
            \node[varNode, label=above:$V_2$] (v2) at (-1, 3 * 2) {};
            \node[varNode, label=above:$\theta_2$] (v4) at (1, 3 * 2) {};
            
            \node[factorSCADA, label=left:$f_{V_2}$] (f6) at (-3, 3 * 2) {};
            \node[factorSCADA, label=right:$f_{P_{12}}$] (f7) at (4, 2 * 2) {};
            
            \draw[edge] (f1) -- (v1);
            \draw[edge] (f4) -- (v3);
            \draw[edge] (f2) -- (v1);
            \draw[edge] (f5) -- (v3);
            \draw[edge] (f5) -- (v1);
            \draw[edge] (f2) -- (v3);
            \draw[edge] (f2) -- (v2);
            \draw[edge] (f2) -- (v4);
            \draw[edge] (f5) -- (v2);
            \draw[edge] (f5) -- (v4);
            
            \draw[edge] (f6) -- (v2);
            \draw[edge] (f7) -- (v1);
            \draw[edge] (f7) -- (v2);
            \draw[edge] (f7) -- (v3);
            \draw[edge] (f7) -- (v4);
            \draw[edge2] (v1) to [out=180,in=180,looseness=1.5] (v2);
            \draw[edge2] (v1) -- (v3);
            \draw[edge2] (v1) -- (v4);
            \draw[edge2] (v2) -- (v3);
            \draw[edge2] (v2) -- (v4);
            \draw[edge2] (v3) to [out=0,in=0,looseness=1.5] (v4);
            
        \end{scope}
    \end{tikzpicture}
    \caption{Factor graph (full-line edges) and the augmented factor graph for a simple two-bus power system. Variable nodes are depicted as circles, and factor nodes are as squares, colored differently to distinguish between phasor and legacy measurements.}
    \label{toyFactorGraph}
    
\end{figure}

Inspired by the application of probabilistic graphical models for the nonlinear power system SE\cite{cosovic2019bpse}, we first transform the bus/branch power system model into the power system's factor graph, a bipartite graph comprised of the factor and variable nodes. We use variable nodes to generate an $s$-dimensional node embedding and the predictions of the state variables, i.e., magnitudes and angles of the bus voltages, ${V}_i$ and ${\theta}_i$, $i=1,\dots,n$. Because variable nodes have no other input features, binary index encoding is added as an input feature to help the GNN model distinguish between similar subgraphs. Factor nodes propagate legacy and phasor measurement values and variances to neighbouring variable nodes via GNN message passing via corresponding node embedding. When creating the factor graph from the bus/branch power system model, each phasor measurement generates two factor nodes, while each legacy measurement generates one factor node. As an example, we consider a simple two-bus power system, in which we placed one voltage phasor measurement on the first bus and one legacy voltage magnitude measurement on the second bus. Additionally, we placed one current phasor measurement and one legacy active power flow measurement on the branch connecting the two nodes. The factor graph of this simple power system consists of the generated factor and variable nodes, connected by full-line edges, as shown in Fig.~\ref{toyFactorGraph}. In contrast to approaches such as \cite{pagnier2021physicsinformed}, in which GNN nodes correspond to state variables only, we find applying GNNs to factor-graph-like topologies more flexible. Since factor nodes can be added or removed from any point in the graph, it is simple to simulate the inclusion of different types and numbers of measurements both on power system buses and branches, such as multiple legacy and phasor measurements on a single branch.

As shown in \cite{kundacina2022state}, augmenting the factor graph by adding direct connections between the variable nodes separated by only one factor node\footnote{Although augmenting the factor graph in the described way no longer makes it bipartite, we will continue to use terms such as augmented factor graph, factor nodes, and variable nodes.} notably improves the model's performance when some portion of measurements are removed from GNN inputs, like when simulating communication failure scenarios. We simulate the measurement data loss by removing the corresponding factor nodes from the factor graph, which can eliminate the connections that physically exist in the power system, reducing the message passing between the nodes. For example, removing both phasor and legacy measurements from the branch in the described two-bus power system divides the factor graph into two isolated parts, disabling the message passing between the pairs of variable nodes. Augmenting the factor graph preserves the physical connection between the variable nodes, improving the message passing in unobservable scenarios. In the two-bus power system example, the augmented factor graph consists of nodes connected by both full and dashed-line edges in Fig.~\ref{toyFactorGraph}.

\begin{figure}[!t]
\centering
\begin{tikzpicture} [scale=0.65, transform shape]

\tikzset{
    varNode/.style={circle,minimum size=6mm,fill=white,draw=black},
    factorVoltage/.style={draw=black,fill=blue!80, minimum size=6mm},
    box/.style={draw, fill=Goldenrod, minimum width=1.5cm, minimum height=0.9cm}}

\begin{scope}[local bounding box=graph]

\node [box]  (layerF0) at (-3, 1.5) {$\layerf$};
\node [box]  (layerV0) at (-3, 0) {$\layerv$};

\node[factorVoltage, label=above:$\mathbf{h_{f}}^{K-1}$] (facReV1) at (0, 1.5) {};
\node[varNode, label=below:$\mathbf{h_{v2}}^{K-1}$] (varImV1) at (0, 0) {};
\node[varNode, label=above:$\mathbf{h_{v}}^{K}$] (varReV1) at (6, 0) {};

\node [box]  (layerV1) at (3, 0) {$\layerv$};
\node [box]  (pred) at (6, -3) {$\pred$};
\node [box, fill=SpringGreen]  (loss) at (3, -3) {Loss};

\draw[-stealth] (-4.25, 1.75) -- (layerF0.west);
\draw[-stealth] (-4.25, 1.5) -- (layerF0.west) node[at start,left]{$...$};
\draw[-stealth] (-4.25, 1.25) -- (layerF0.west);

\draw[-stealth] (-4.25, 0.25) -- (layerV0.west);
\draw[-stealth] (-4.25, 0) -- (layerV0.west) node[at start,left]{$...$};
\draw[-stealth] (-4.25, -0.25) -- (layerV0.west);

\draw[-stealth] (layerF0.east) -- (facReV1.west);
	
\draw[-stealth] (layerV0.east) -- (varImV1.west);
	
\draw[-stealth] (varImV1.east) -- (layerV1.west);
	
\draw[-stealth] (0.25, -2.5) -- (layerV1.west)
	node[at start,left]{$...$} node[very near end,left]{$...$};
	
\draw[-stealth] (facReV1.east) -- (layerV1.west);
	
\draw[-stealth] (layerV1.east) -- (varReV1.west);
	
\draw[-stealth] (varReV1.south) -- (pred.north);
	
\draw[-stealth] (pred.west) -- (loss.east)
	node[midway,above]{$output$};
	
\draw[-stealth] (3, -2.1) -- (loss.north)
	node[at start,above]{$label$};

\end{scope}

\end{tikzpicture}

\hfil

\caption{The high-level computational graph displays the final message passing iteration along with prediction and loss calculation for the variable node $v$. Yellow rectangles represent trainable functions implemented as feed-forward neural networks.}
    \label{computationalGraph}
\end{figure}
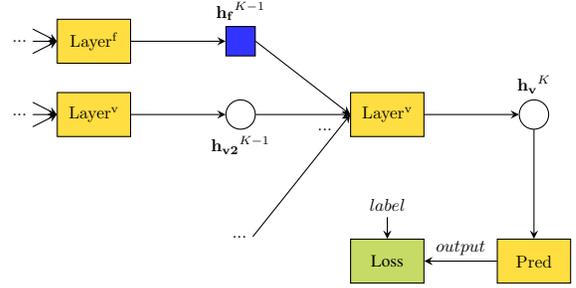

We propose a GNN architecture specialised for working on heterogeneous augmented factor graphs, consisting of two types of GNN layers, $\layerf(\cdot): \mathbb {R}^{\textrm{deg}(f)  \cdot s} \mapsto \mathbb {R}^{s}$ used for updating embeddings of factor nodes $f$, and $\layerv(\cdot): \mathbb {R}^{\textrm{deg}(v) \cdot s} \mapsto \mathbb {R}^{s}$ for variable nodes $v$, with different sets of trainable parameters. In addition to variable-to-factor and factor-to-variable node message functions, with the augmentation, we introduce new variable-to-variable node messages and we model them with separate trainable message function $\textrm{Message}\textsuperscript{v$\rightarrow$v}(\cdot): \mathbb {R}^{2s} \mapsto \mathbb {R}^{u}$. For all of the message functions we used two-layer feed-forward neural networks, single layer neural networks for the update functions and the aggregation function based on the attention mechanism from graph attention networks \cite{velickovic2018graph}. State variables predictions $\mathbf{x^{pred}}$ are generated by feeding the final variable node embeddings $\mathbf h^K$ into the additional two-layer neural network $\pred(\cdot): \mathbb {R}^{s} \mapsto \mathbb {R}$. Equations \eqref{embeddingds_and_predictions} describe the recursive neighbourhood aggregation and the state variable prediction processes:
\begin{equation}
    \begin{gathered}
        \mathbf {h_v}^k = \layerv(\{\mathbf {h_i}^{k-1} | i \in \mathcal{N}_v\})\\
        \mathbf {h_f}^k = \layerf(\{\mathbf {h_i}^{k-1} | i \in \mathcal{N}_f\})\\
        {x_v}^{pred} = \pred(\mathbf {h_v}^K)\\
        k \in \{1,\dots,K\},
    \end{gathered}
    \label{embeddingds_and_predictions}
\end{equation}
where $\mathcal{N}_v$ and $\mathcal{N}_f$ denote the $1$-hop neighbourhoods of variable and factor nodes $v$ and $f$. We use the mean squared difference between state variable predictions and labels $\mathbf{x^{label}}$ as a training loss function, calculated over the entire mini-batch of graphs:
\begin{equation} \label{loss_function}
    \begin{gathered}
        L(\theta^\textrm{GNN}) = \frac{1}{2nB} \sum_{i=1}^{2nB}({{x_i}^{pred}} - {{x_i}^{label}})^2,
    \end{gathered}
\end{equation}
where $2n$ is the total number of variable nodes in a graph, $B$ is the mini-batch size, and $\theta^\textrm{GNN}$ represents all the trainable parameters. Fig.~\ref{computationalGraph} displays a high-level computational graph of the final message passing iteration, state variable prediction, and the loss function calculation for the part of the augmented factor graph given in Fig.~\ref{toyFactorGraph}.

The power system augmented factor graph has bounded node degree, prediction for a single node requires only information from the node's $K$-hop neighbourhood, and the number of GNN layers is small. Analogously to \cite{kundacina2022robust}, we can conclude that the computational complexity of the proposed model's inference for a single state variable is constant, which implies that the overall complexity of the proposed GNN-based SE is $\mathcal{O}(n)$. The implementation of the GNN model in large-scale networks can be further improved by distributing the inference computation among local processors in the power system, avoiding the communication delays between the measurement devices and the central processing unit in the centralised SE implementation. For arbitrary $K$, measurements required for the single complex bus voltage inference are located within the $\lceil K/2 \rceil$-hop neighbourhood of the corresponding power system bus.

\section{Numerical Results and Discussion}

In this section, we describe the GNN model's training process and test the trained model on various examples to validate its accuracy, and its robustness under measurement data loss due to communication failure and cyber attacks in the form of malicious data injections. The described GNN model for augmented factor graphs is implemented using the IGNNITION library \cite{pujolperich2021ignnition} and trained for 100 epochs in all experiments using the following hyperparameters: 64 elements in the node embedding vector, 32 graphs in a mini-batch, four GNN layers, ReLU activation functions, Adam optimiser, and a learning rate of $4\times 10^{-4}$. We conducted separate training experiments for IEEE 30 and IEEE 118-bus test cases, for which we generated a training set containing 10000 samples and validation and test sets containing 100 samples each. Each sample is created by randomly sampling the active and reactive power injections and solving the power flow problem. Measurement values are created by adding Gaussian noise to the power flow solutions, and the nonlinear SE is solved by GN to label the input measurement set in each sample. We used a Gaussian noise variance of $10^{-5}$ for phasor measurements, $10^{-3}$ for bus voltage magnitude and active and reactive power flow legacy measurements, and $10^{-1}$ for active and reactive injection legacy measurements.

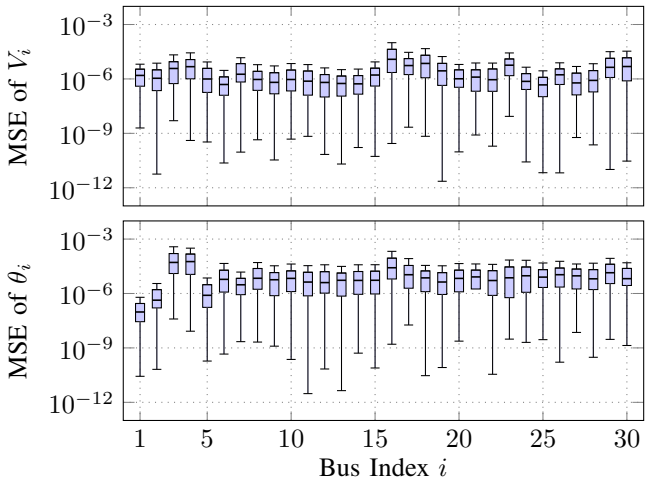
\begin{figure}[ht]
	\centering
	\begin{tikzpicture}
	
	\begin{groupplot}[group style={group size=1 by 2, vertical sep=0.2cm}, ]
	\nextgroupplot[box plot width=0.6mm, ymode=log,
        	ylabel={MSE of $V_i$},
  	        grid=major,   	
  	        yscale=0.8,
  	        xmin=0, xmax=31, ymin=1e-13, ymax = 1e-2,	
            xtick={1,5,10,15,20,25,30}, 
            xticklabels={,,},
  	        ytick={1e-12, 1e-9, 1e-6, 1e-3},
  	        width=8.5cm,height=4.9cm,
  	        legend style={draw=black,fill=white,legend cell align=left,font=\tiny, at={(0.01,0.82)},anchor=west}
  	        ]
	
	        \boxplot [
                forget plot, fill=blue!20,
                box plot whisker bottom index=1,
                box plot whisker top index=5,
                box plot box bottom index=2,
                box plot box top index=4,
                box plot median index=3] {./plots/box_plot_magnitude_mse.txt};

	    \nextgroupplot[box plot width=0.6mm, ymode=log,
    	    xlabel={Bus Index $i$},
        	ylabel={MSE of ${\theta}_i$},
  	        grid=major, 
  	        yscale=0.8,
  	        xmin=0, xmax=31, ymin=1e-13, ymax = 1e-2,
            xtick={1,5,10,15,20,25,30}, 
  	        ytick={1e-12, 1e-9, 1e-6, 1e-3},
  	        width=8.5cm,height=4.9cm,
  	        legend style={draw=black,fill=white,legend cell align=left,font=\tiny, at={(0.01,0.82)},anchor=west}]
	
	        \boxplot [
                forget plot, fill=blue!20,
                box plot whisker bottom index=1,
                box plot whisker top index=5,
                box plot box bottom index=2,
                box plot box top index=4,
                box plot median index=3] {./plots/box_plot_angle_mse.txt};  
    \end{groupplot}        
	\end{tikzpicture}
	\caption{The test set MSE between the predictions and the labels per each bus for voltage magnitudes and angles in the IEEE 30-bus test case.}
	\label{plot2}
\end{figure}

For the IEEE 30-bus test case, we placed $100$ legacy measurements and three PMUs (i.e., three bus voltage phasors and eight branch current phasors) in each sample, resulting in $2.03$ measurement redundancy. The trained model performed well on the test set, with the average test set mean square error of $1.233\times 10^{-5}$ between predictions and ground truth labels; the average test set MSE for voltage magnitudes of $5.221\times 10^{-6}$; the average test set MSE for voltage angles of $1.944\times 10^{-5}$. Fig.~\ref{plot2} shows the average test MSE per each bus, where the upper plot corresponds to voltage magnitudes and the lower one to voltage angles.

For the IEEE 118-bus test case, we placed $500$ legacy measurements and seven PMUs (i.e., seven bus voltage phasors and $26$ branch current phasors) in each sample, resulting in $2.39$ measurement redundancy. The average test set mean square error equals $2.038\times 10^{-5}$, with the average test set MSE for voltage magnitudes of $1.572\times 10^{-5}$ and the average test set MSE for voltage angles of $2.505\times 10^{-5}$. Based on the insights from both experiments, we can conclude that the proposed GNN model is a good approximator of the nonlinear SE solved by GN.

\subsection{Robustness to Loss of Input Data}
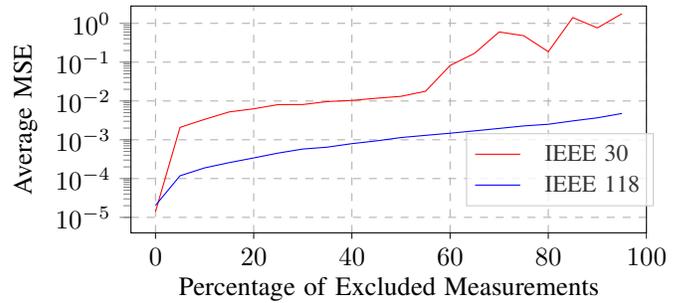
\begin{figure}[htbp]
    \centering
    \pgfplotstableread[col sep = comma,]{plots/test_result_columns.csv}\testResultColumns
    \begin{tikzpicture}
    
    \begin{axis}[
    yscale=0.53,
    ymode=log,
    legend cell align={left},
    legend columns=1,
    legend style={
      fill opacity=0.8,
      font=\small,
      draw opacity=1,
      text opacity=1,
      at={(0.65,0.01)},
      anchor=north west,
      draw=white!80!black
    },
    tick align=outside,
    tick pos=left,
    x grid style={white!69.0196078431373!black},
    xlabel={Percentage of Excluded Measurements},
    xmin=-5, xmax=100, ymin=0.000004, ymax=2.8,
    xtick style={color=black},
    xlabel style={yshift=-4pt},
    y grid style={white!69.0196078431373!black},
    ylabel={Average MSE},
    ytick={0.00001, 0.0001, 0.001, 0.01, 0.1, 1},
    scaled y ticks=false,
    xmajorgrids=true,
    ymajorgrids=true,
    grid style=dashed
    ]
    \addplot [red]
    table [x={x-axis}, y={averageMSEsIEEE30}]{\testResultColumns};
    \addlegendentry{IEEE 30}
    \addplot [blue]
    table [x={x-axis}, y={averageMSEsIEEE118}]{\testResultColumns};
    \addlegendentry{IEEE 118}
    
    \end{axis}
    
    \end{tikzpicture}
    \caption{Average MSEs of test sets created by randomly excluding measurements.}
    \label{msesForBothModels}
\vspace{-4mm}    
\end{figure}

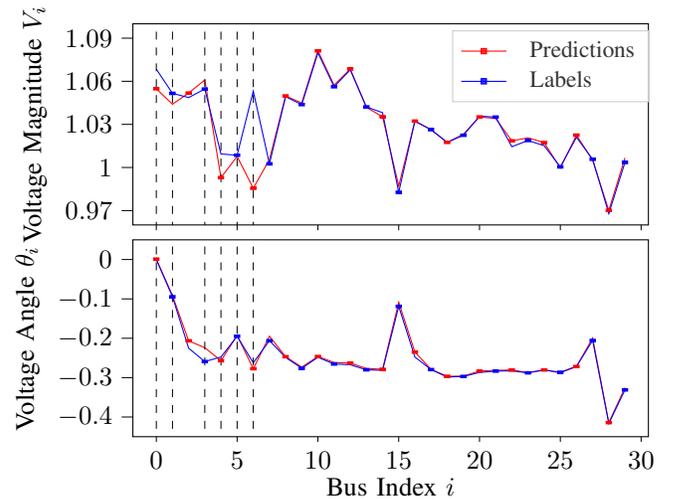
\begin{figure}[htbp]
    \centering
    \pgfplotstableread[col sep = comma,]{plots/PredictionsPerNodeMeasFrom2Nodesxcluded.csv}\CSVPredVSObservdTwoPMUExcludedEdgesVarNodes
    
        \begin{tikzpicture}
        
        \begin{groupplot}[group style={group size=1 by 2, vertical sep=0.2cm}, ]
        \nextgroupplot[
        yscale=0.47,
        legend cell align={left},
        legend columns=1,
        legend style={
          fill opacity=0.8,
          font=\small,
          draw opacity=1,
          text opacity=1,
          at={(0.81,25.14)},
          anchor=south,
          draw=white!80!black,
          nodes={scale=0.98, transform shape}
        },
        scaled x ticks=manual:{}{\pgfmathparse{#1}},
        tick align=outside,
        tick pos=left,
        x grid style={white!69.0196078431373!black},
        xmin=-1.45, xmax=30.45,
        xtick style={color=black},
        xticklabels={},
        y grid style={white!69.0196078431373!black},
        ylabel={\text{Voltage Magnitude} ${V}_i$},
        ymin=0.96, ymax=1.1,
        ytick style={color=black},
        ytick={0.97, 1, 1.03, 1.06, 1.09},
        ]
        \addplot [red, mark=square*, mark options={scale=0.5}, mark repeat=2,mark phase=1]
        table [x expr=\coordindex, y={predictionsRE}]{\CSVPredVSObservdTwoPMUExcludedEdgesVarNodes};
        \addlegendentry{Predictions}
        
        \addplot [blue, mark=square*, mark options={scale=0.5}, mark repeat=2,mark phase=2]
        table [x expr=\coordindex, y={true_valuesRE}]{\CSVPredVSObservdTwoPMUExcludedEdgesVarNodes};
        \addlegendentry{Labels}
        
        
        
        \addplot [very thin, black, dashed, forget plot]
        table {%
        1 -2.0
        1 2.0
        };
        \addplot [very thin, black, dashed, forget plot]
        table {%
        0 -2.0
        0 2.0
        };
        \addplot [very thin, black, dashed, forget plot]
        table {%
        6 -2.0
        6 2.0
        };
        \addplot [very thin, black, dashed, forget plot]
        table {%
        5 -2.0
        5 2.0
        };
        \addplot [very thin, black, dashed, forget plot]
        table {%
        4 -2.0
        4 2.0
        };
        \addplot [very thin, black, dashed, forget plot]
        table {%
        3 -2.0
        3 2.0
        };

        \nextgroupplot[
        yscale=0.46,
        tick align=outside,
        tick pos=left,
        x grid style={white!69.0196078431373!black},
        xlabel={Bus Index $i$},
        xmin=-1.45, xmax=30.45,
        xtick style={color=black},
        xlabel style={yshift=-4pt},
        y grid style={white!69.0196078431373!black},
        ylabel={\text{Voltage Angle} ${\theta}_i$},
        ymin=-0.45, ymax=0.05,
        ytick style={color=black},
        ]
        
        \addplot [red, mark=square*, mark options={scale=0.5}, mark repeat=2,mark phase=1]
        table [x expr=\coordindex, y={predictionsIM}]{\CSVPredVSObservdTwoPMUExcludedEdgesVarNodes};
        
        \addplot [blue, mark=square*, mark options={scale=0.5}, mark repeat=2,mark phase=2]
        table [x expr=\coordindex, y={true_valuesIM}]{\CSVPredVSObservdTwoPMUExcludedEdgesVarNodes};
        
        
        
        \addplot [very thin, black, dashed, forget plot]
        table {%
        1 -2.0
        1 2.0
        };
        \addplot [very thin, black, dashed, forget plot]
        table {%
        0 -2.0
        0 2.0
        };
        \addplot [very thin, black, dashed, forget plot]
        table {%
        6 -2.0
        6 2.0
        };
        \addplot [very thin, black, dashed, forget plot]
        table {%
        5 -2.0
        5 2.0
        };
        \addplot [very thin, black, dashed, forget plot]
        table {%
        4 -2.0
        4 2.0
        };
        \addplot [very thin, black, dashed, forget plot]
        table {%
        3 -2.0
        3 2.0
        };
        
        \end{groupplot}
        
        \end{tikzpicture}

    \caption{GNN predictions and labels for one test example with all measurements connected to two neighbouring buses removed. Dashed lines indicate the buses in the $1$-hop neighbourhood of the excluded measurements.}
    \label{PredictionsPerNodeMeasFrom2Nodesxcluded}
\end{figure}

Next, we observe predictions of the GNN models previously trained on IEEE 30 and IEEE 118-bus test data when exposed to the loss of input data caused by communication failures or measurement device malfunctions. We simulate the described cases by randomly removing a percentage of all input measurements, ranging from $0\%$ to $95\%$ with a step of $5\%$. We create 20 test sets per IEEE test case, each containing samples with the same percentage of excluded measurements, and show the average test set MSEs in Fig.~\ref{msesForBothModels}. Proposed GNN models yields predictions in all examples, with an expected growing trend in MSE as the number of excluded measurements increases. In comparison, the GN method could not provide a solution for many examples due to underdetermined and ill-conditioned systems of nonlinear SE equations. A possible explanation for significantly lower MSEs for the IEEE 118-test case in these scenarios is that it contains a greater variety of subgraphs for GNN training.
To investigate the GNN predictions further, we create a test set by excluding five measurements connected to the two directly connected power system buses from each test sample, resulting in the average test set MSE of $1.488\times 10^{-4}$. Fig.~\ref{PredictionsPerNodeMeasFrom2Nodesxcluded} shows the results for one test sample, where vertical dashed lines correspond to the buses in the $1$-hop neighbourhood of the excluded measurements. We can observe that the deviation from the ground truth values manifests mainly in the vicinity of the excluded measurements, not affecting the prediction accuracy in the rest of the power system.

 \subsection{Behavior Under Malicious Data Injections}
\begin{figure}[htbp]
    \centering
    \pgfplotstableread[col sep = comma,]{plots/PredictionsPerNodeWithAttacks.csv}\PredictionsPerNodeWithAttacks
    
        \begin{tikzpicture}
        
        \begin{groupplot}[group style={group size=1 by 2, vertical sep=0.2cm}, ]
        \nextgroupplot[
        yscale=0.54,
        legend cell align={left},
        legend columns=1,
        legend style={
          fill opacity=0.8,
          font=\small,
          draw opacity=1,
          text opacity=1,
          at={(0.79,0.99)},
          anchor=south,
          draw=white!80!black
        },
        scaled x ticks=manual:{}{\pgfmathparse{#1}},
        tick align=outside,
        tick pos=left,
        x grid style={white!69.0196078431373!black},
        xmin=-1.45, xmax=30.45,
        xtick style={color=black},
        xticklabels={},
        y grid style={white!69.0196078431373!black},
        ylabel={Voltage Magnitude ${V}_i$},
        ymin=0.94, ymax=1.147,
        ytick style={color=black},
        xmajorgrids=true,
        ymajorgrids=true,
        ytick={0.96, 1, 1.04, 1.08, 1.12},
        ]
        \addplot [red, mark=square*, mark options={scale=0.5}, mark repeat=3,mark phase=1]
        table [x expr=\coordindex, y={predictionsRE}]{\PredictionsPerNodeWithAttacks};
        \addlegendentry{Predictions}

        
        \addplot [black, mark=square*, mark options={scale=0.5}, mark repeat=3,mark phase=3]
        table [x expr=\coordindex, y={WLS_SE_with_attack_RE}]{\PredictionsPerNodeWithAttacks};
        \addlegendentry{GN based SE}
        
        \addplot [blue, mark=square*, mark options={scale=0.5}, mark repeat=3,mark phase=2]
        table [x expr=\coordindex, y={true_valuesRE}]{\PredictionsPerNodeWithAttacks};
        \addlegendentry{Ground truth}
        
        
        \nextgroupplot[
        yscale=0.42,
        tick align=outside,
        tick pos=left,
        x grid style={white!69.0196078431373!black},
        xlabel={Bus Index $i$},
        xlabel style={yshift=-4pt},
        xmin=-1.45, xmax=30.45,
        xtick style={color=black},
        y grid style={white!69.0196078431373!black},
        ylabel={Voltage Angle ${\theta}_i$},
        ymin=-0.44, ymax=0.04,
        ytick style={color=black},
        xmajorgrids=true,
        ymajorgrids=true,
        ytick={-0.4, -0.3, -0.2, -0.1, 0},
        ]
        
        \addplot [red, mark=square*, mark options={scale=0.5}, mark repeat=3,mark phase=1]
        table [x expr=\coordindex, y={predictionsIM}]{\PredictionsPerNodeWithAttacks};
        
        \addplot [blue, mark=square*, mark options={scale=0.5}, mark repeat=3,mark phase=2]
        table [x expr=\coordindex, y={true_valuesIM}]{\PredictionsPerNodeWithAttacks};
        
                
        \addplot [black, mark=square*, mark options={scale=0.5}, mark repeat=3,mark phase=3]
        table [x expr=\coordindex, y={WLS_SE_with_attack_IM}]{\PredictionsPerNodeWithAttacks};
        
        
        \end{groupplot}
        
        \end{tikzpicture}
    \caption{GNN predictions and GN based SE solutions for one test example with corrupted input data.}
    \label{PredictionsPerNodeCyberAttacks}
\end{figure}
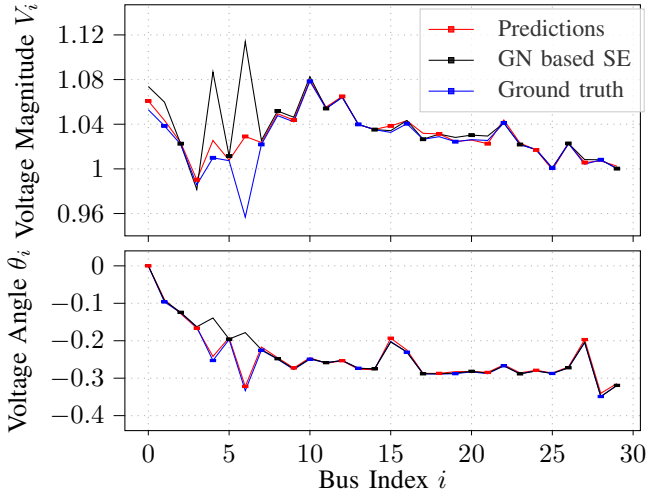

We examine the robustness of the proposed GNN model to malicious data injection type of cyber attacks by randomly altering the values of five neighbouring measurements in each test sample. We compare the proposed GNN model's predictions with the solutions of the GN method and the ground truth values obtained using the GN method applied on the uncorrupted measurement data. The GNN model demonstrated an order of magnitude better performance than the GN method, with the average test set MSEs $1.281\times 10^{-4}$ and $1.034\times 10^{-3}$, respectively. Fig.~\ref{PredictionsPerNodeCyberAttacks} depicts the comparison of the state variable predictions under corrupted input data for one example from the test set.

\section{Conclusion}
In this paper, we introduced a method for nonlinear SE considering legacy and phasor measurements based on the GNN model specialised for operating on augmented power system factor graphs. The method avoids the problems that traditional nonlinear SE solvers face, such as numerical instabilities and convergence time depending on the state variable initialisation. Additional benefits of the proposed GNN model are linear computational complexity regarding the number of state variables during the inference phase and the possibility of distributing the inference computation across multiple processing units. By testing the GNN on power systems of various sizes, we observed the prediction accuracy in the normal operating states of the power system and the sensitivity when encountering false data injection cyber attacks and input data loss due to communication irregularities.

Since the proposed GNN model generates predictions even for underdetermined SE systems of equations, it could be applied to highly unobservable distribution power systems. Another application of the proposed model could be the fast and accurate initialisation of the nonlinear SE solver, resulting in a hybrid approach that is both model-based and data-driven.

\section{Acknowledgment}
This paper has received funding from the European Union's Horizon 2020 research and innovation programmes under Grant Agreement numbers 856967 and 871518.

\bibliographystyle{IEEEtran}
\bibliography{cite}

\end{document}